%% file: root.tex
\documentclass[letterpaper, 10 pt, conference]{ieeeconf}  

\IEEEoverridecommandlockouts                              

\usepackage{graphicx} 
\usepackage{amsmath} 
\usepackage{subfigure}
\usepackage[]{algorithmic}
\usepackage{algorithm}
\usepackage{siunitx}

\title{\LARGE \bf
MobiKa - Low-Cost Mobile Robot for Human-Robot Interaction
}

\author{Florenz Graf$^{1}$, \c{C}a\u{g}atay Odaba\c{s}{\i}$^{1}$, Theo Jacobs$^{1}$, Birgit Graf$^{1}$ and Thomas F{\"o}disch$^{2}$
\thanks{$^{1}$The authors are researchers with the Robot and Assistive Systems Department, Fraunhofer IPA, 70569 Stuttgart, Germany. {\tt\small <first name>.<last name>@ipa.fraunhofer.de } }%
\thanks{$^{2}$Thomas F{\"o}disch is innovation manager of elderly care, BruderhausDiakonie, 72762 Reutlingen, Germany
        {\tt\small thomas.foedisch@bruderhausdiakonie.de}}
}

\begin{document}

\maketitle
\thispagestyle{empty}
\pagestyle{empty}

\begin{abstract}
One way to allow elderly people to stay longer in their homes is to use of service robots to support them with everyday tasks. With this goal, we design, develop and evaluate a low-cost mobile robot to communicate with elderly people. The main idea is to create an affordable communication assistant robot which is optimized for multimodal Human-Robot Interaction (HRI). Our robot can navigate autonomously through dynamic environments using a new algorithm to calculate poses for approaching persons. The robot was tested in a real life scenario in an elderly care home.
\end{abstract}


\input{sections/intro}
\input{sections/related_work}
\input{sections/robot_design}

\input{sections/approaching_human}
\input{sections/results}
\input{sections/conclusion}

\input{sections/acknowledgement}

                     
\bibliographystyle{plain}             
\bibliography{ref.bib}

\end{document}

%% file: sections/intro.tex
\section{INTRODUCTION}
\label{sec:intro}

Due to demographic change in the coming years, the number of elderly people will increase in most industrialized countries.  Robot technology can help these people to live self-determined and independent in their homes as long as possible and to reduce the need for ambulant or stationary care, e.g. by providing means of communication, detecting anomalies and emergencies, guiding people and fetching objects. Service robots can also support other people with reduced mobility such as rehabilitation patients.

Private households are highly dynamic enviroments which are primarily designed for humans. Therefore a mobile robot has to cope with narrow passages and needs a design that supports safe HRI. To be affordable, robots need to be availiable at low cost, but offer at the same time considerable functionality.

\begin{figure}[ht!]
	\centering
	\includegraphics[width=0.48\textwidth ]{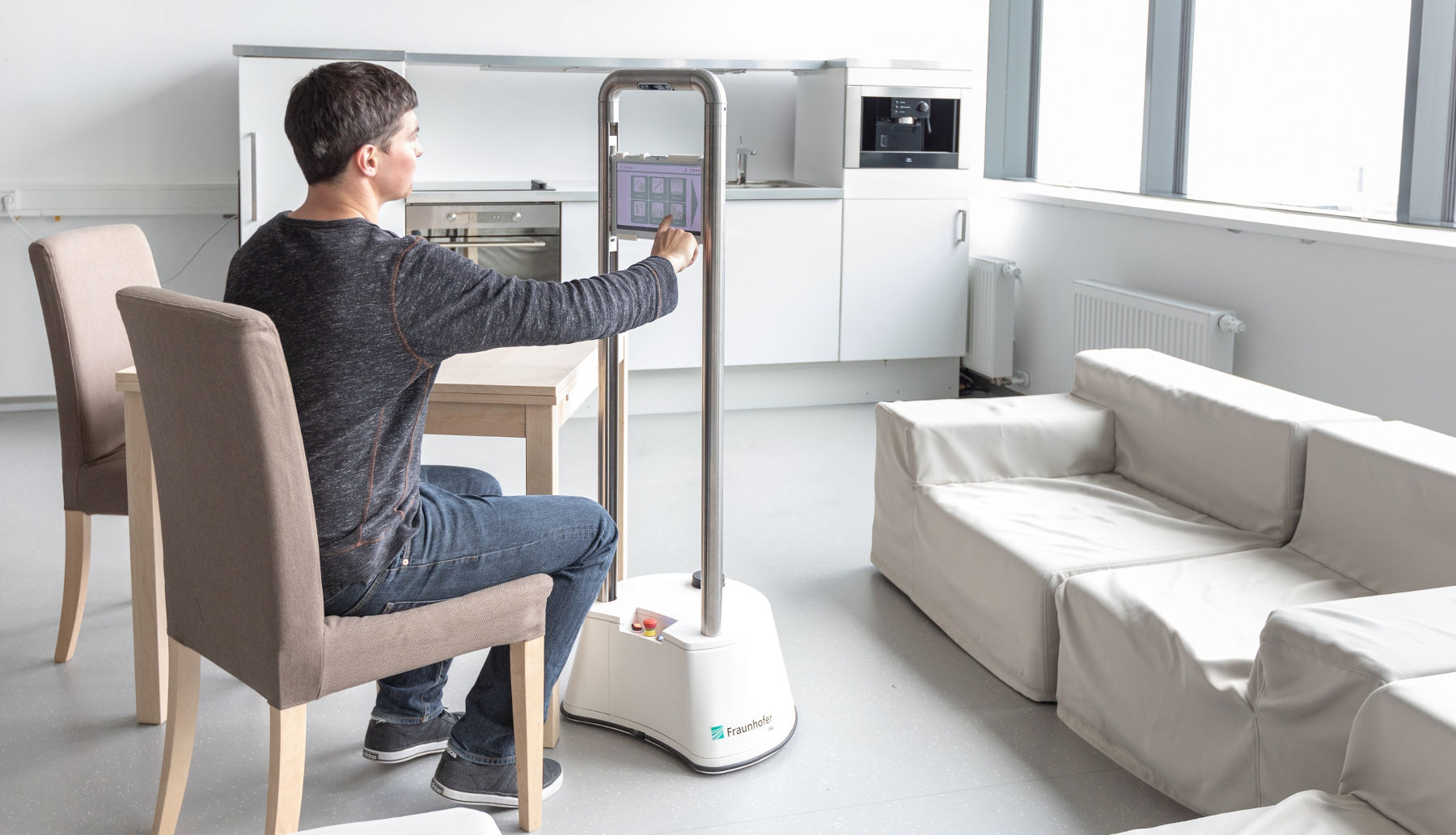}
	\label{fig:mobika-interacting}
	\caption{MobiKa is interacting with a sitting human}
\end{figure}

In this paper, we introduce MobiKa (Mobile Communication Assistant) depicted in Fig. \ref{fig:mobika-interacting}. Our vision is to solve the aformentioned issues by developing an affordable multi-purpose mobile service robot focusing on communication. MobiKa can navigate autonomously within a pre-mapped environment. For Human-Robot Interaction, MobiKa is able to approach robustly the user. MobiKa is easy to use, even for non-technical users. With the use of functional hardware design and modular software architecture, we provide a highly adaptable robot platform.
 
This paper is organized as follows. In Section~\ref{sec:relwork}, the related work is introduced to the reader. Our robot's hardware and software designs are explained in detail in Section~\ref{sec:robotdesign}. Next, the proposed approaching humans is described in Section~\ref{sec:approaching} and the experiments and their evaluation are introduced in Section~\ref{sec:results}. Lastly, the paper is briefly concluded in Section~\ref{sec:conclusion} with a short outlook.

%% file: sections/related_work.tex
\section{RELATED WORK}
\label{sec:relwork}

In health care, service robots have the potential to help humans in different ways such as physical, emotional, social and cognitive. A typical service robot includes a user-interface, several sensors, e.g. cameras and laser scanner, a mobile base and sometimes a pair of arms. Care-O-bot \cite{kittmann2015let}, Homemate \cite{zhao2014octree} and PR2, Toyota HSR \cite{Yamamoto:2018:HSR:3214907.3233972} and TIAGo \cite{tiago2019} are good examples of fully capable service robots. They can fetch and carry objects for people, navigate autonomously indoor, perceive a person with their cameras and approach them for interaction. Since the arms should be safe and able to carry payload, the technical requirements are high. The arms make the robot multifunctional, but also complex and expensive, hence today not affortable by the end users. In contrast to such multifunctional robots, there are also more specialized robots with fewer functionalities. One example is Pepper \cite{pandey2018mass, ahn2018development} that has arms just for carrying out gestures to reinforce the expression of user-interaction. Other robots like the Robotic Service Assistant \cite{baumgarten2016robotic} or SMOOTH demonstrator \cite{juel2018smooth} do not have arms but are still able to carry out physical tasks. The Robotic Service Assistant serves people drinks by driving to the user and handing out drinks. The SMOOTH demonstrator assists elderly people by transporting objects. Additionally, some robots cannot physically support persons. These robots are consisting of auditive I/O and visual I/O e.g. Kompai \cite{kompai2010}, SCITOS \cite{scitos2007} or RP-VITA \cite{rpvita2019} to interact with people. However, they can observe people and communicate with them to remind them to take pills, to socialize, or to monitor the health. ElliQ \cite{elliq2019} is an example of a stationary companion which allows people to do phone calls and play cognitive games. There are also social robots like iSocioBot \cite{tan2018isociobot, tan2015designing}. This robot socializes with people by observing them and creating some facial expressions and speech. Since it is a research platform, its size is enormous compared to Zenbo and Buddy \cite{milliez2018buddy}. Another example of specialized robots is MobiNa \cite{mobina2015} - a low-cost robot with navigation for emergency assistance.

For the Human-Robot Interaction, robots need to know how to approach a person by using human-aware navigation to maintain human comfort \cite{Ramirez2016, KRUSE20131726, lee2017}. The process consists of different stages \cite{Satake2009}: Finding person for interaction, interacting in public distance, initiating conversation in social distance. 

%% file: sections/robot_design.tex
\section{ROBOT DESIGN}
\label{sec:robotdesign}

MobiKa is designed as a mobile communication assistant. While designing the robot, the main idea was to create an affordable system optimized for Human-Robot Interaction. Therefore we chose a functional design which helps us to reduce the price and illustrates, that the robot's capabilities are far away from that of a human. To make it affordable, we based the design on open-source software and low-cost hardware. 

The goal of the development was to cover functions such as:
\begin{itemize}
	\item General communication tasks via multi-modal interfaces (speech and visual)
	\item  Entertainment functions (games and services on display, activating the user)
	\item  Reaction to users falling; connection with stationary sensors and networks to allow
	detection of medical emergencies and contacting an external service provider 
	(robot guides to fallen person)
	\item  Reminding of appointments and taking medication
	\item Telepresence and telemedicine
	\item  Simple transport tasks (user places objects on the robot)
	\item  Guiding persons
	\item  Open infrastructure to third-party apps, e.g., for medical services
	
\end{itemize}

With these functions, MobiKa can support elderly persons to stay independent and live longer in their homes. It can also assist rehabilitation patients so that they can return to their normal
every-day routine earlier. 

\subsection{Hardware Design}\label{sec:hw}
\begin{figure}
	\subfigure[MobiKa as an emergency assistant]{\includegraphics[width=0.24\textwidth]{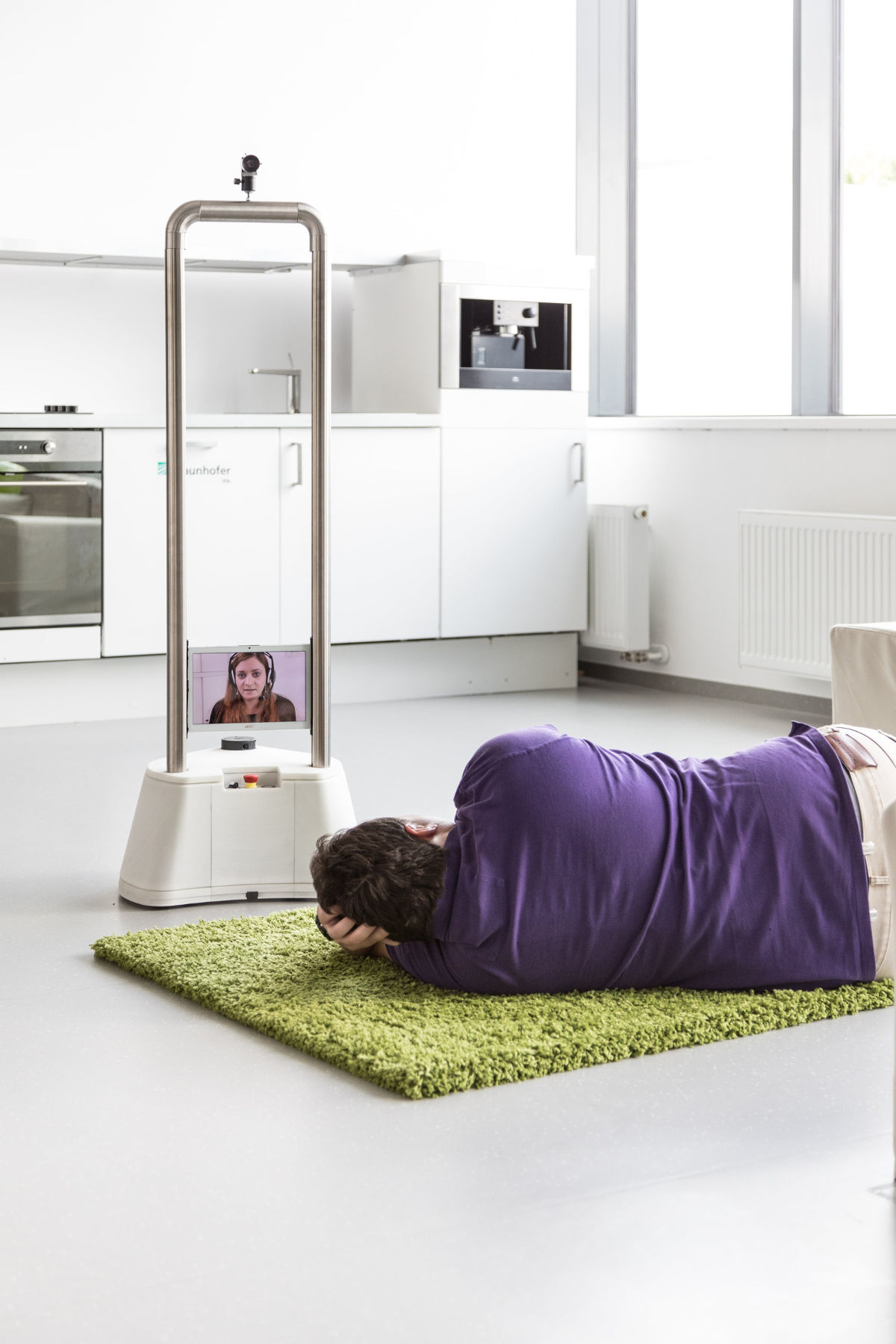}}
	\subfigure[MobiKa as a social assistant]{\includegraphics[width=0.24\textwidth]{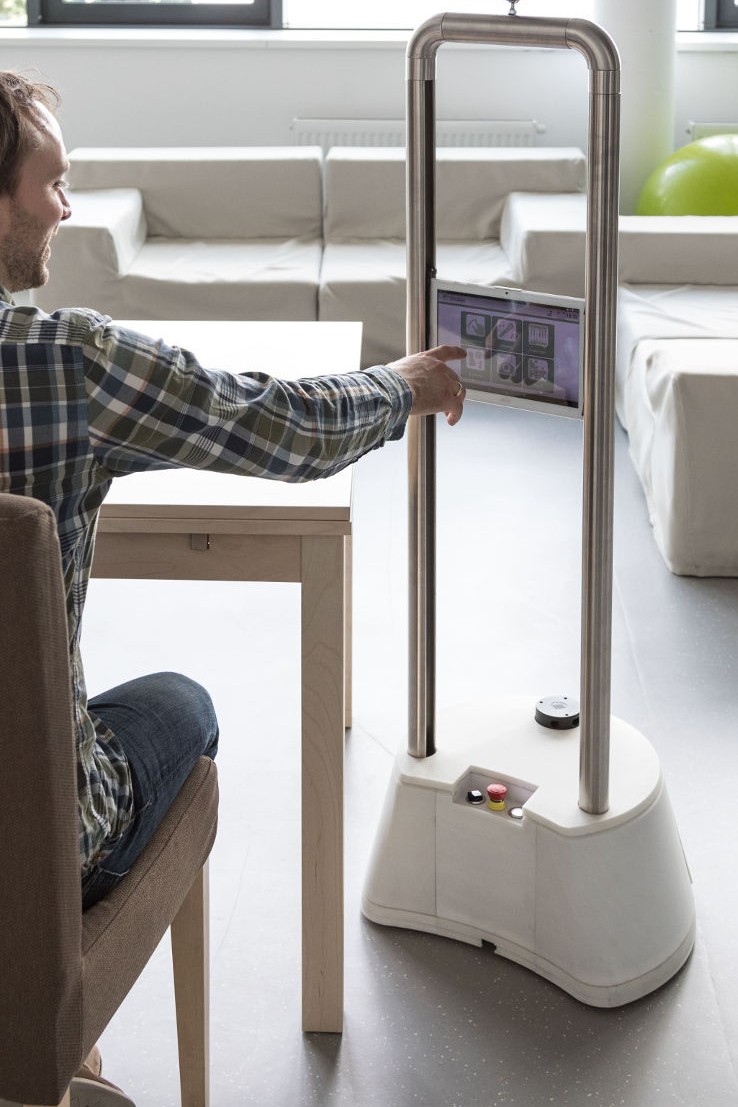}}
	
	\label{fig:fdfhh}
	\caption{Different use cases of MobiKa. Thanks to its adjustable tablet height MobiKa can interact with the people while they are standing, sitting and laying (e.g. after a fall).}
	\label{fig:mobika_adaptive}
\end{figure}

MobiKa is built on a compact mobile base in which the main components reside. The dimensions of the robot were derived from the intended user interaction; to interact with standing persons, MobiKa needs a minimum height of $\SI{1.1}{\metre}$. To also allow interaction with sitting and lying persons (Fig. \ref{fig:mobika_adaptive}), the screen needs to be adjustable in height. Therefore a belt-driven linear axis was designed, that allows adjusting an Android tablet to the pose of the user. Length and width and also mass were kept as low as possible, which required to concentrate the mass close to the ground to maintain stability during travel.

The design is based on low-cost components. While the differential drive and also the tablet axis are formed by simple 
DC kit motors and motor controllers, the $\SI{24}{\V}$ battery originates from an e-bike. For the tablet axis, belts and pullies
from 3D printer supply were chosen. Selecting the processing unit was also fundamental since it should be low-cost, 
energy-efficient but performant. That’s why we picked an octa-core computing device Odroid XU4 with eMMC flash storage. 
We found that it is capable of running the software that the robot needs with minimal power consumption. Other components
include DC-DC converters and a Wi-Fi bridge. As it became clear very soon, that the outer shell of the robot is also costly 
(e.g., when 3D-printed or molded in small quantities), a design was chosen which limits covers to the mobile base, while the structure to support the tablet is made of metal tubes. To keep the vertical axis simple, all cabling between tablet and robot base was avoided. The tablet only connects to charging contacts in its lower end position. Communication is solved via Wi-Fi.

MobiKa's sensors consist of a low-cost LIDAR sensor, which provides distance data in a $\SI{360}{\degree}$ angle in a horizontal plane and a 3D sensor, which looks downward in a $\SI{45}{\degree}$ angle along the front of the robot and allows detecting persons, tables and small obstacles on the ground. In future, the camera will be attached on an additional axis which will enable it to look forward (e.g., to recognize faces of a standing person) and also backward (e.g., for docking a battery charger). For safety reasons, the robot is also equipped with a small bumper to detect collisions. 

\subsection{Software Design}

\subsubsection{Software Structure} 

MobiKa needs several independent software components interacting with each other in real time. ROS (Robot Operating System) \cite{quigley2009ros} running on Ubuntu 16.04 is responsible for the communication of these components. Thanks to ROS, we could create a decentralized and modular software system meaning that none of the packages depends on a central application other than the ROS master. This is crucial for complex robotic systems since if there is a failure in one of the software components or hardware drivers, we can directly diagnose the failed components and fix the issue without affecting the other parts of the robot. This is also the case for updating the individual software components.  

\subsubsection{Virtual Model}

The software should be aware of the links and joints of the robot. To support this,  we create the virtual model of the robot using URDF (Unified Robot Description Format), which is an XML based robot description format. Thanks to URDF along with CAD design, we can successfully visualize the robot in our software.

\subsubsection{Navigation}

The navigation software stack is one of the most critical parts of MobiKa's software. It enables the robot to navigate to the person safely inside known environment. The environment is represented by a 2D gridmap that is created initially by the robot using the open-source GMapping package \cite{grisetti6gmapping}. During navigation, a simultaneos updated costmap inflates obstacles from the gridmap and from the sensor data to limit the operating area for collision avoidance. Due to the issue that the laser scanner scans the environment only horizontally at its mounting hight, obstacles at other heights are not visible. This is dangerous because the robot can collide with tables or other objects that are not fully detectable by the laser scanner. The 3D sensor at the top of the robot solves this issue by projecting a 3D point cloud onto the gound plane as a virtual scan. Using the virtual scan as additional input of the costmap helps that the robot is able to navigate safely without any collision. 
The navigation makes use of an EKF (Extended Kalman Filter) to localize the robot inside the gridmap. The EKF is part of the Fraunhofer IPA navigation stack, which uses the wheel odometry and laser scanner data as input. When we launch the robot, the last known robot pose is set as initial pose. Afterwards the wheel odometry updates the pose incrementally. In case of association between map features and laser scanner data the pose will be corrected. This is necessary to compensate odometry drift.

%% file: sections/approaching_human.tex
\section{APPROACHING A HUMAN}
\label{sec:approaching}

Before a robot approaches a human of interest, the robot needs to know where the human is and where to move in relation to the human based on the environment model.The pose of the human is provided to the robot by an external pose detection system.

Defining the robot goal pose by a fixed offset to the human is not robust because obstacles could make the goal unreachable. Like Human-Human-Interaction, there is a variable set of possible poses for HRI. The set of possible poses highly increases the probability of finding a reachable robot goal. This is addressed by the new method we developed. All possible poses are describing an area around the human which is delimited by the operating range of the user (Section \ref{sec:searcharea}).
During the approach to the human, the robot needs to continuously update the whole calculation of the goal area (Section IV-B) because of the dynamics in the environment (e.g., the user is moving) and the limited field of view. The robot may not always perceive the goal area.
Our implementation solves the issues by using a dynamic recalculation of a set of goal positions based on the grid map of the robot environment. The grid-based calculation dynamically makes an efficient rating of multiple goal cells in the search area. This allows us to continuously update this procedure during the approach of the user and avoids the previously mentioned problems. The temporary best-rated cell becomes the goal pose (Section \ref{sec:best}), extended by the orientation from this cell to the center of the human.

\subsection{Defining Search Area} \label{sec:searcharea}

The search area (Fig. \ref{fig:searchsteps}a) is the area that humans can reach with their arms to interact with the robot. The operating area of a human $P(x,y,\varphi)$ is limited. Minimum radius $r_{min}$ and maximum radius $r_{max}$ around the human limit the operating distances. Moreover, angle constraints limit the operating orientation range of the human. For each valid radius, the software calculates a circle around the human inside the costmap and check the angle conditions using algorithm 1.
 
\begin{algorithm}
	\label{alg:searcharea}
	\caption{Defining goal grid cells for HRI}
	\textsc{DefiningSearchArea($gridmap,P(x,y,\varphi)$)}
	
	\begin{algorithmic}[!h]
		\STATE{Set seedpoint at $P(x,y,\varphi)$} 
		\FOR{$r_{min}$ \TO $r_{max}$} 
		\STATE{calculate circle in gridmap:}
		\FOR{$cell$ in $circle$} 
		\IF{angle conditions are TRUE} \STATE{ADD $cell$ to $S$} \ENDIF 
		\ENDFOR
		\ENDFOR
		\STATE{\textbf{Return} $S$}
		
	\end{algorithmic}
\end{algorithm}

Flexible parameters are defining the angle constraints. If the human is standing, the robot tries to approach from the front into the unidirectional search area, defined by $\alpha_1$. If the human is sitting, the robot tries to approach from the sides into the bidirectional search area, defined by $\alpha_2$. All suitable cells $i$ are stored inside a container $S$.

\subsection{Calculating Costs} \label{sec:calc}

All positions inside the search area are suitable for human-robot interaction but may be unreachable for the robot. The idea behind this calculation is to quantify the suitability by costs and to reveal unreachable cells inside the container.
The calculation uses the latest costmap of the robot. The suitability rates the costmap and the path planning for the robot; it also rates the distance and angle error for the human. The sum of those four influences indicates the overall quality of each cell.

\subsubsection{Costmap Cost} \label{sec:cm}
The first calculation step checks the value of the costmap at the position $C(x_i, y_i)$ of each cell inside the container (see Fig. \ref{fig:searchsteps}b). The value of the costmap multiplied by an influence factor $m_{cm}$ assigns the costmap cost to the cells. Further calculations are ignoring occupied cells. In real environments, this step efficiently reduces the number of cells.

\begin{equation*}
c_{i,cm} = C(x_i, y_i) \cdot m_{cm}
\end{equation*}

\subsubsection{Path Planning Cost} \label{sec:path}

Even if the costmap is not occupied, the robot may not be able to find a path to this cell, i.e., the cell is unreachable. This calculation step approves and rates the path planning of all cells in one shot. The path search starts at the robot position exploring all neighbors using the breadth-first search (BFS) algorithm of Lee [9], which is very efficient. The search ends after reaching all goal cells or if the exploring depth is disproportionate to the distance. Removing all unreached cells from the container reduces the further calculation (see Fig. \ref{fig:searchsteps}c).
In contrast to common path planning algorithms like the A* which provides the optimum for just one goal, the Lee-Algorithm provides the optimum for all goal cells in one shot. The overall length of the path between the robot and the goal cells $l_i$ multiplied by an influence factor $m_{path}$ indicates the cost of the remaining cells.

\begin{equation*}
c_{i,path} = l_i \cdot m_{path}
\end{equation*}

\subsubsection{Distance Cost} \label{sec:dist}
For every human the optimal operation distance to the robot is different. The distance cost $c_{i,dist}$ describes this influence by assigning a radial cost function starting at the center of the human, e.g., linear increasing by radius $r_i$ as shown in Fig.~\ref{fig:searchsteps}d. This personalization helps to prefer the optimal user interaction distance.

\begin{equation*}
c_{i,dist} = r_i \cdot m_{dist}
\end{equation*}

\subsubsection{Angle Error Cost} \label{sec:angle}
The angle error cost $c_{i, angle}$ is similar to the previous cost but focus on the robot orientation in relation to the human. The assumption is that the mean angle $\alpha_{mean}$ of the search area(s) is the optimal orientation for HRI. For every cell, the angle difference between the mean angle $\alpha_{mean}$ and cell angle $\alpha_{i}$, multiplied by an influence factor $m_{angle}$ defines the angle cost error (see Fig.~\ref{fig:searchsteps}e).

\begin{equation*}
c_{i,angle} = |\alpha_{mean}-\alpha_{i}| \cdot m_{angle}
\end{equation*}

\subsection{Finding the Best Pose} \label{sec:best}
\begin{figure}[!h]

    \subfigure[Definition Search Area]{ \includegraphics[width=0.18\textwidth]{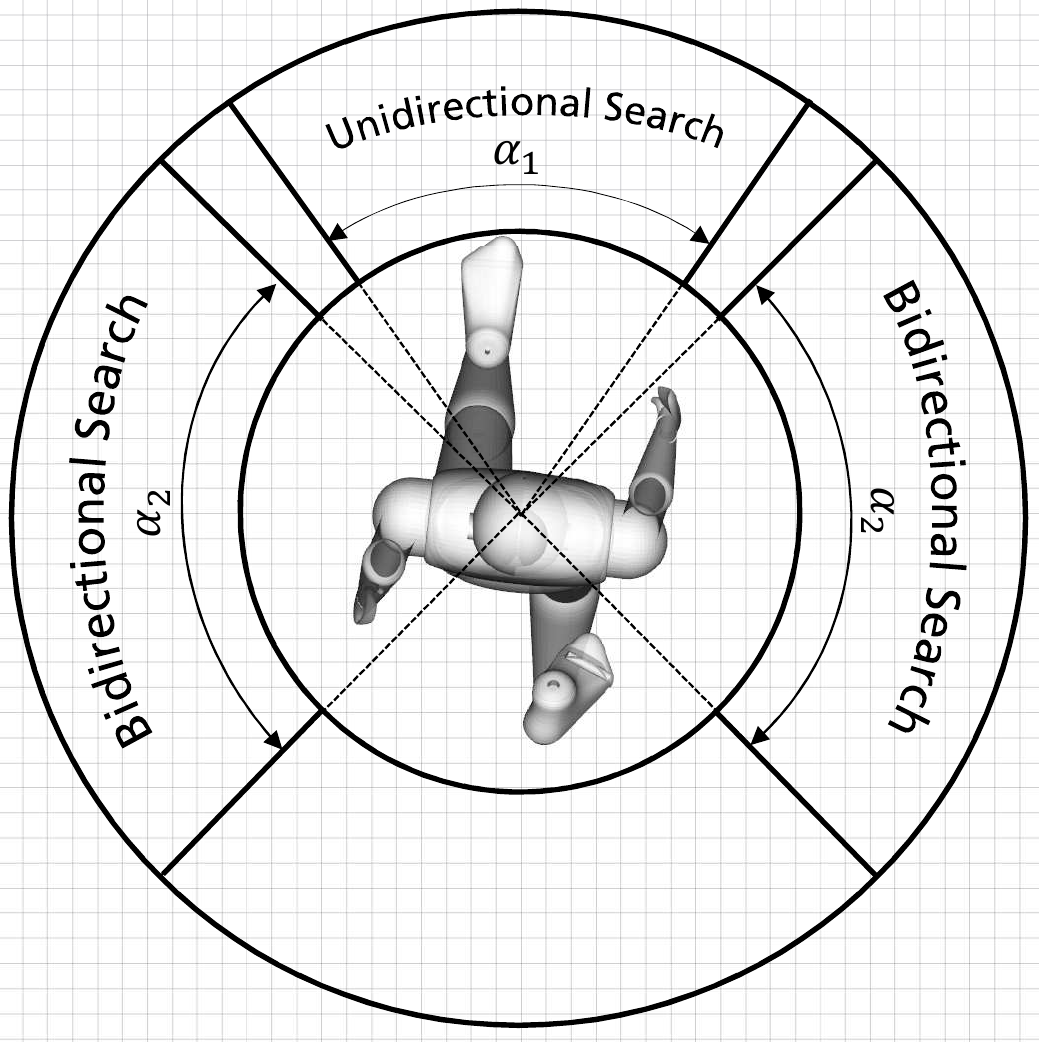}}
	\subfigure[Costmap Cell Weights]{\includegraphics[width=0.24\textwidth]{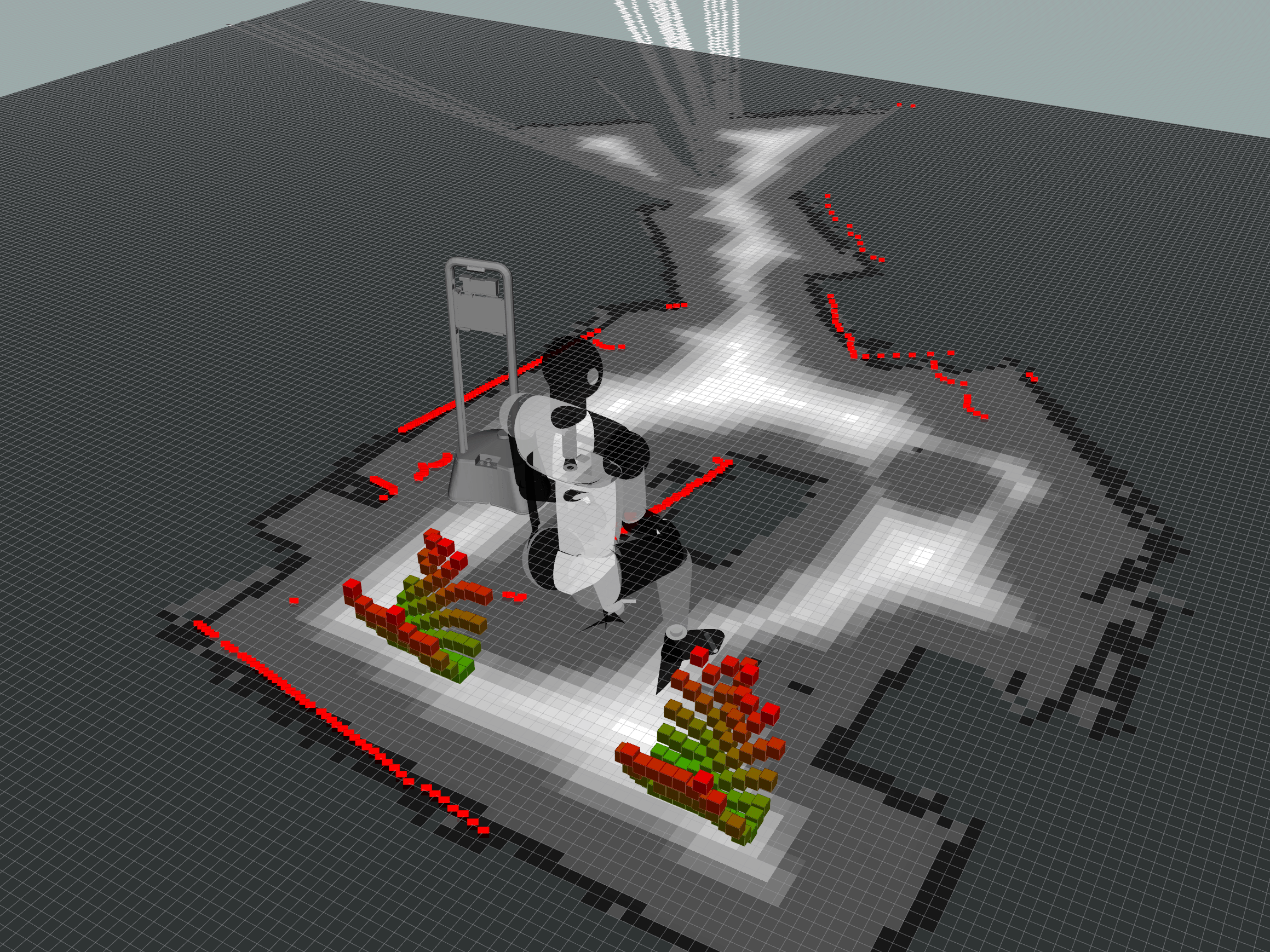}}
    \subfigure[Path Planning Cell Weights]{\includegraphics[width=0.24\textwidth]{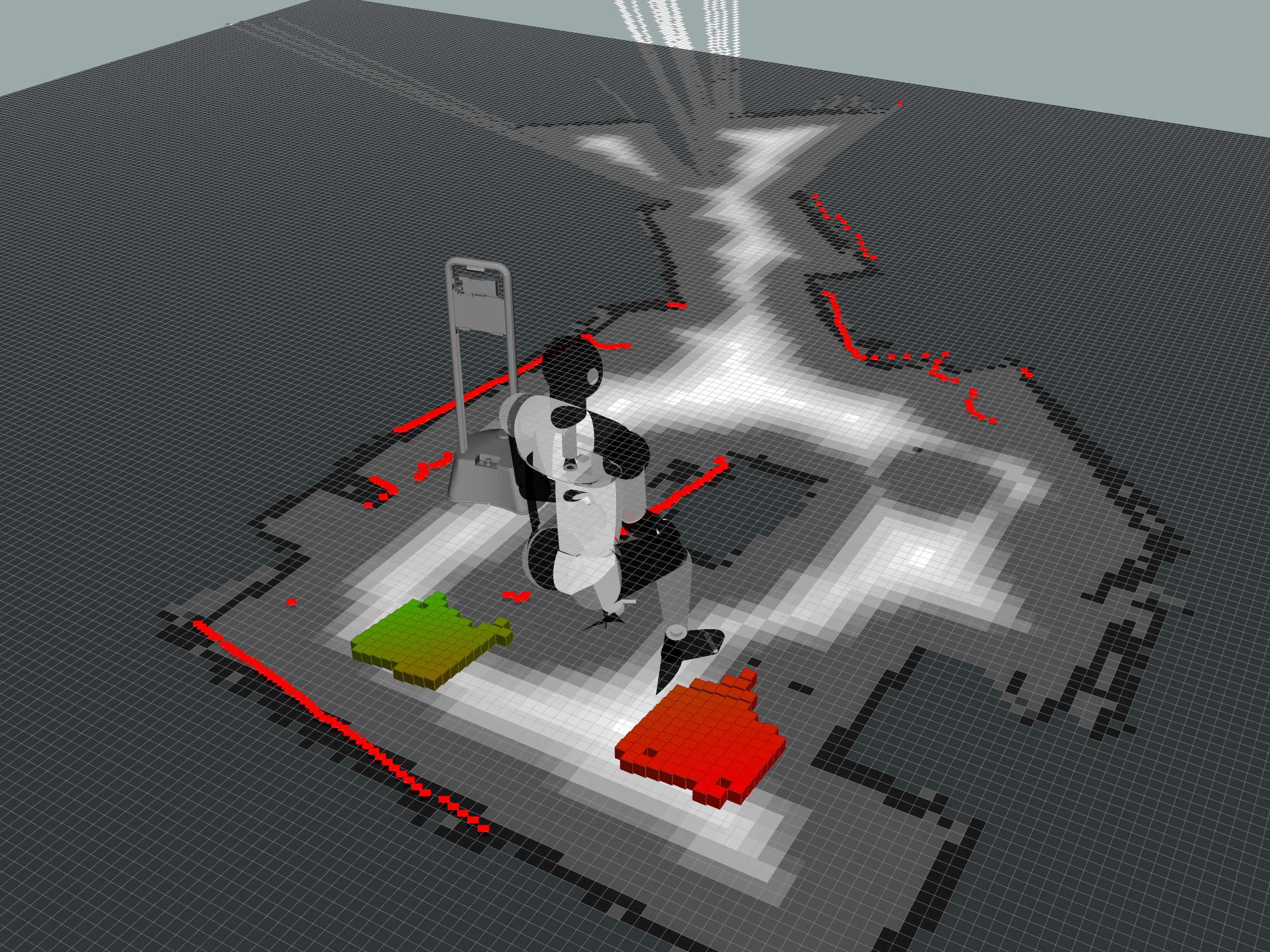}}
\subfigure[Distance Cell Weights]{\includegraphics[width=0.24\textwidth]{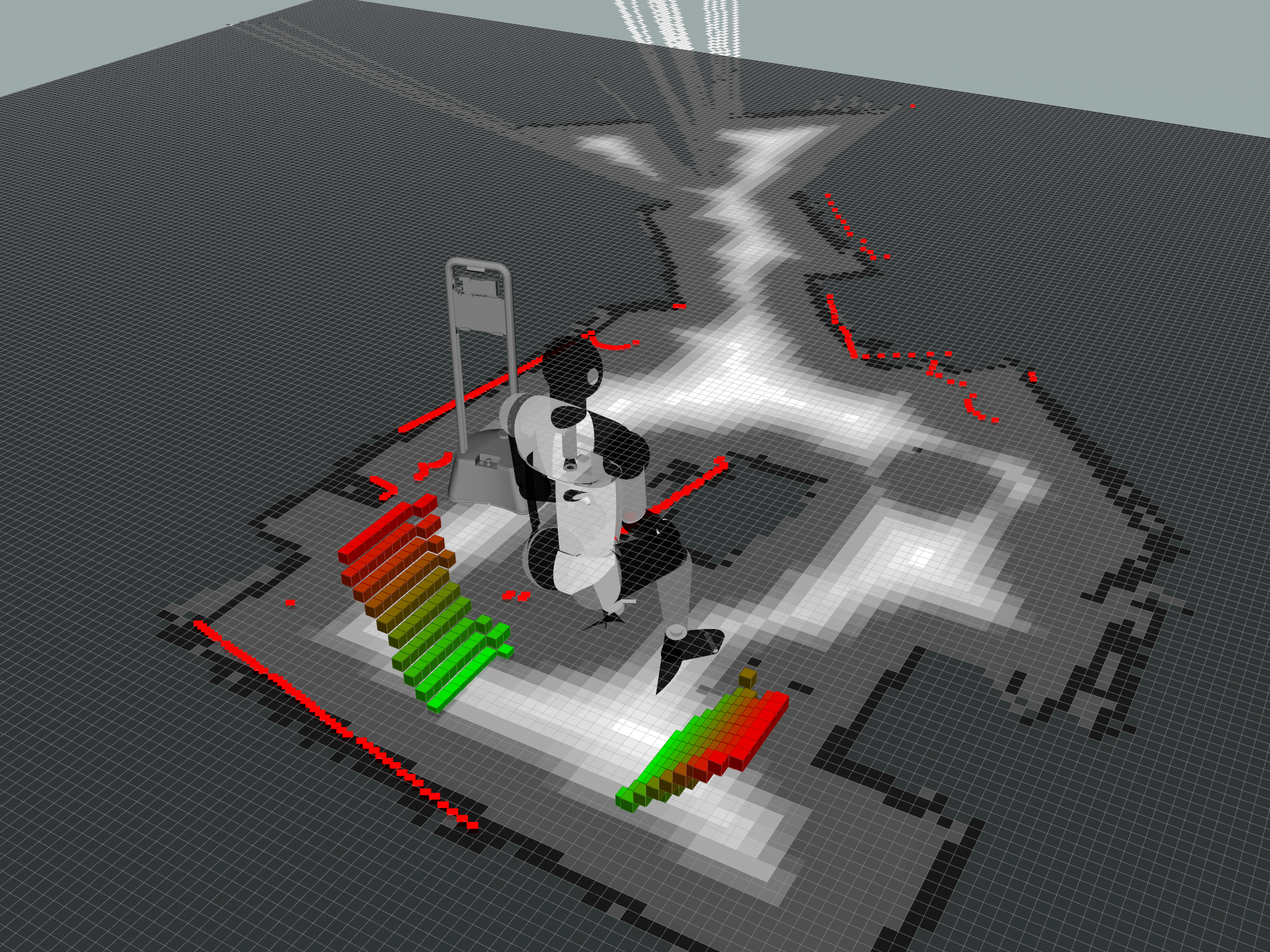}}
\subfigure[Radius Error Cell Weights]{\includegraphics[width=0.24\textwidth]{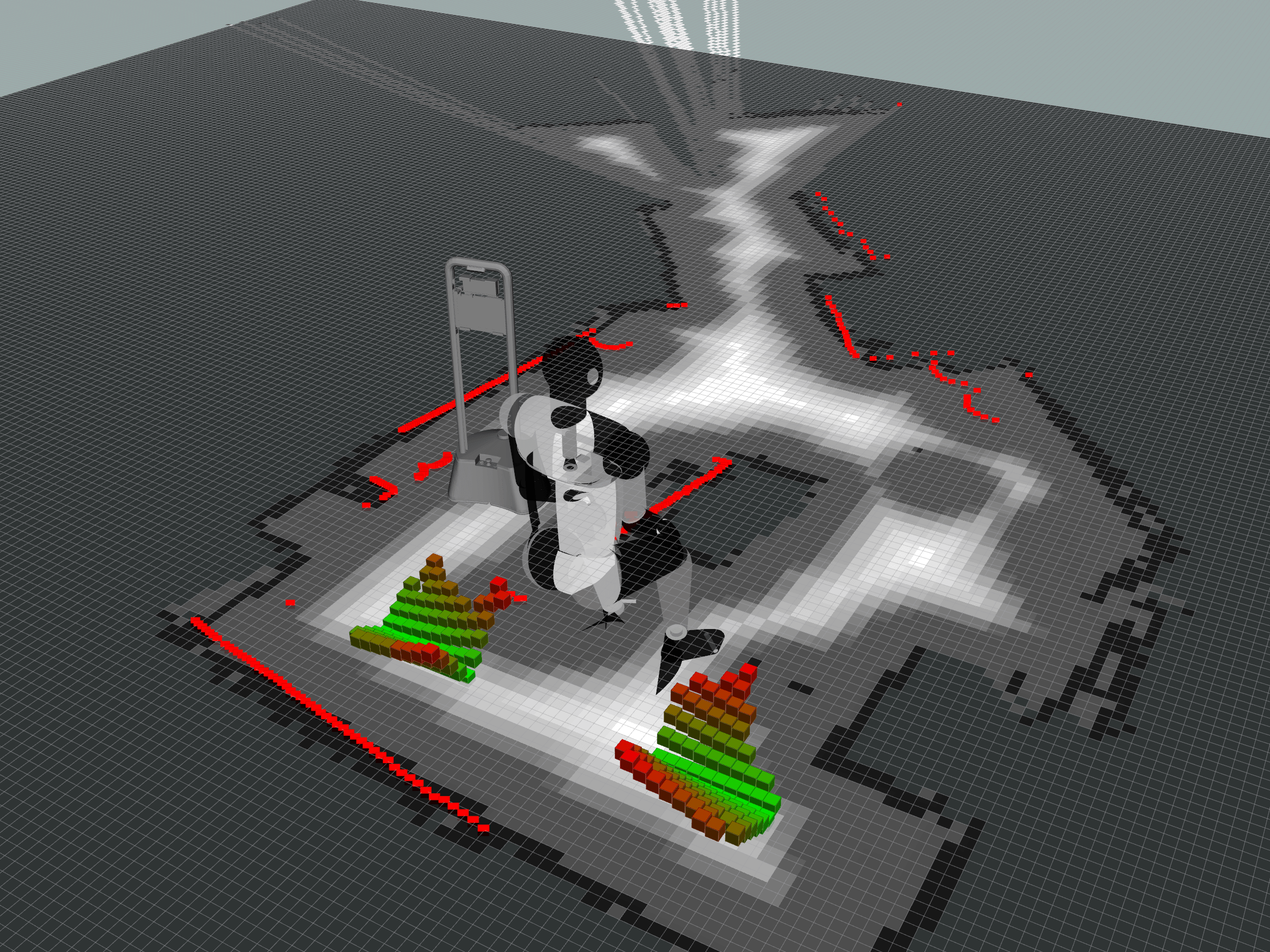}}
    \subfigure[Overall Cell Weights]{\includegraphics[width=0.24\textwidth]{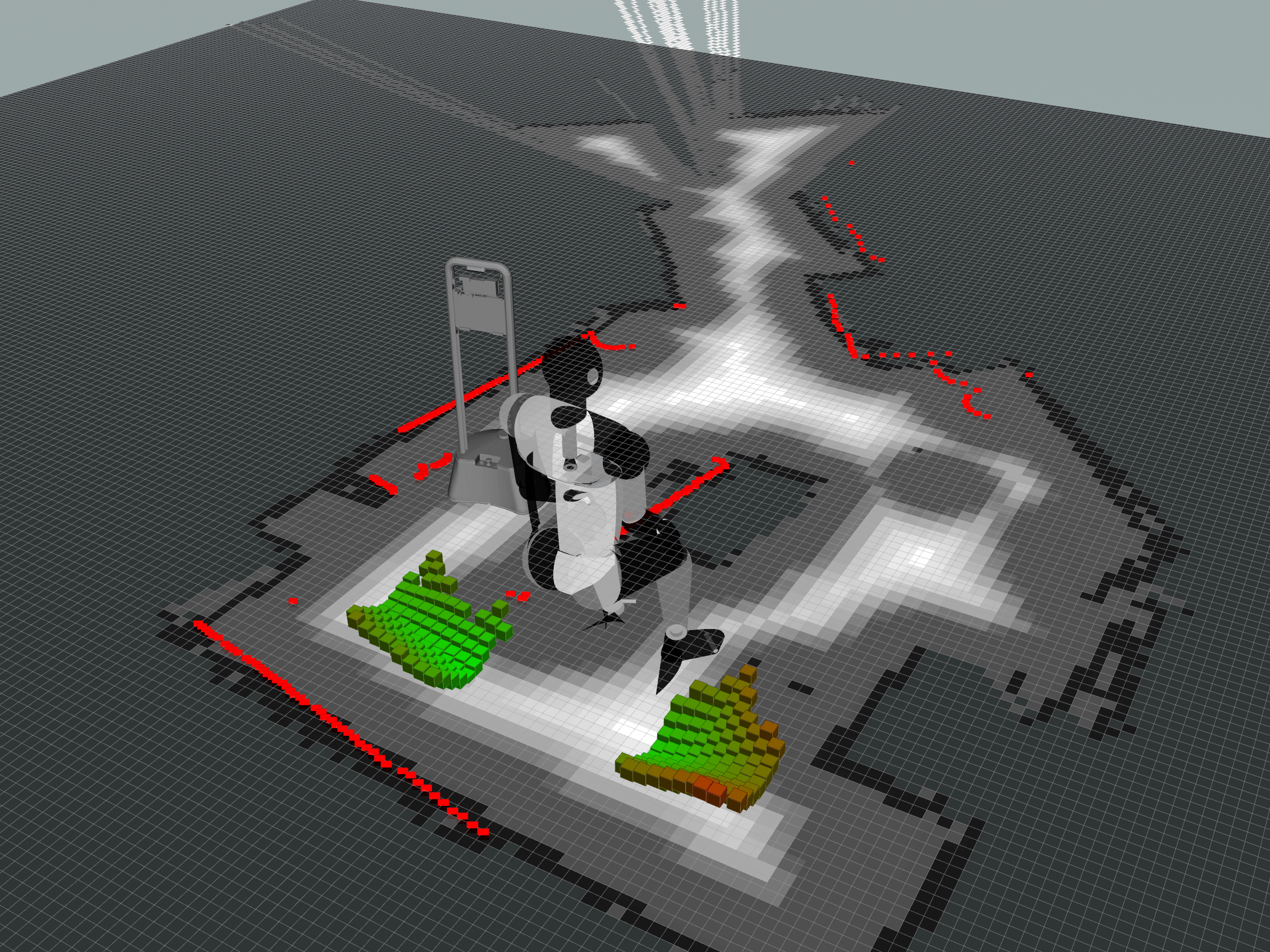}}
   \caption{Different stages of costs for HRI visualized on the gridmap}
   \label{fig:searchsteps} 
\end{figure}

The four influences assign costs to every reachable cell of the goal area of the robot. The sum of all four costs represents the overall weight of every cell.
The overall weight is adjustable by adapting the multiplied influence factors of each cost. The best robot position for HRI is cell $c_{best}$, which has the lowest overall costs i.e. minimizing the sum of costs (see Fig.~\ref{fig:searchsteps}f)

\begin{equation*}
c_{best} = \min_{\forall i \in S} (c_{i,cm} + c_{i,path} + c_{i,dist} + c_{i,angle})
\end{equation*}

For HRI the robot has to look in the direction of the human, which defines the robot orientation $\alpha_r$. The 2D-pose $g$ is the robot goal that is send to move the base. The best robot pose updates permanently during approaching. This recalculation allows adapting dynamically to environment.

\begin{equation*}
g = (c_{best}(x), c_{best}(y), \alpha_r)
\end{equation*}

with $ \alpha_r = \alpha_{h} - \arctan\left(\dfrac{y_{cm} - y_{h}}{x_{cm} - x_{h}}\right) + \pi$

\bigbreak

%% file: sections/results.tex
\section{Evaluation}
\label{sec:results} 

\subsection{General Evaluation} \label{sec:general}
 
After the robot was set up, the basic functionality was verified in our lab. Using the 2D laser scanner in combination with a 3D sensor MobiKa is able to navigate safely in a known environment with the Fraunhofer IPA navigation software. It was further checked that MobiKa's height-adjustable tablet can adapt to standing, sitting and even lying people (see Fig. \ref{fig:mobika_adaptive}). During the publicly funded project EmAsIn \cite{emasin}, the partners could successfully connect their software components (e.g. speech recognition and graphical user interface) to the flexible software framework of MobiKa. Customized apps as well as third-party apps for entertainment were successfully tested. At typical usage, the battery of MobiKa can power the robot for more than eight hours without charging. In general it could be verified that the low-cost components used for MobiKa provide the necessary functionality and robustness.
 
\subsection{Analyzing Approaching To Human in Lab}\label{sec:app}

For the evaluation of the approaching strategy of the robot, we analyzed the final HRI poses within a lab at Fraunhofer IPA (see Fig. 4). After mapping the environment, we set the poses of five human’s static on the map so that errors orginating from an imprecise camera detection could be excluded. The robot had to approach the two people on the sofa unidirectionally from the front ($\alpha_{mean} = \SI{0}{\degree}$). Moreover, the robot had to approach the three people sitting at the table bidirectionally from the sides ($\alpha_{mean} =\SI{60}{\degree}$), even if the best orientation would be to approach from the front \cite{Dautenhahn2006, Kheng2007}. We set the minimum search radius $r_{min}$ to $\SI{0.45}{\metre}$ and $r_{max}$ to $\SI{0.9}{\metre}$ to stay above the intimate distance and still inside the working space of the human arms \cite{hall1966hidden, KRUSE20131726}. The angle $\alpha_{1} = \alpha_{2}$ is set to $\SI{90}{\degree}$.

We defined an approaching sequence for all persons within a simple state machine. The robot navigated ten rounds autonomously where the robot approached all people successfully. During the approaching, the robot goal pose $g$ updated at $\SI{2}{Hz}$. The final robot poses, as well as the poses of the people, are visualized in Fig. 4. Here the small arrows indicate the final robot pose and the bigger arrows indicate the person poses.
Fig. \ref{fig:results} shows the final robot distance and orientation in relation to the center of the humans head split by unidirectional and bidirectional search for 50 poses. The distance reaches from $\SI{0.57}{\metre}$ to $\SI{0.92}{\metre}$ while the orientation $\alpha$ for the unidirectional search was $\SI{0}{\degree} - \SI{10}{\degree}$ and the orientation for the directional was $\SI{79}{\degree} - \SI{103}{\degree}$. In comparison to the table poses, the distances of the sofa poses are higher (see Fig. \ref{fig:results}) because of the user legs and the sofa itself which avoided the robot to approach closer. Moreover, the robot always chose the shortest path, indicated by the side of approaching the persons at the table (see Fig. 4).

\begin{figure}[ht!]
            \centering
            \includegraphics[width=0.4\textwidth]{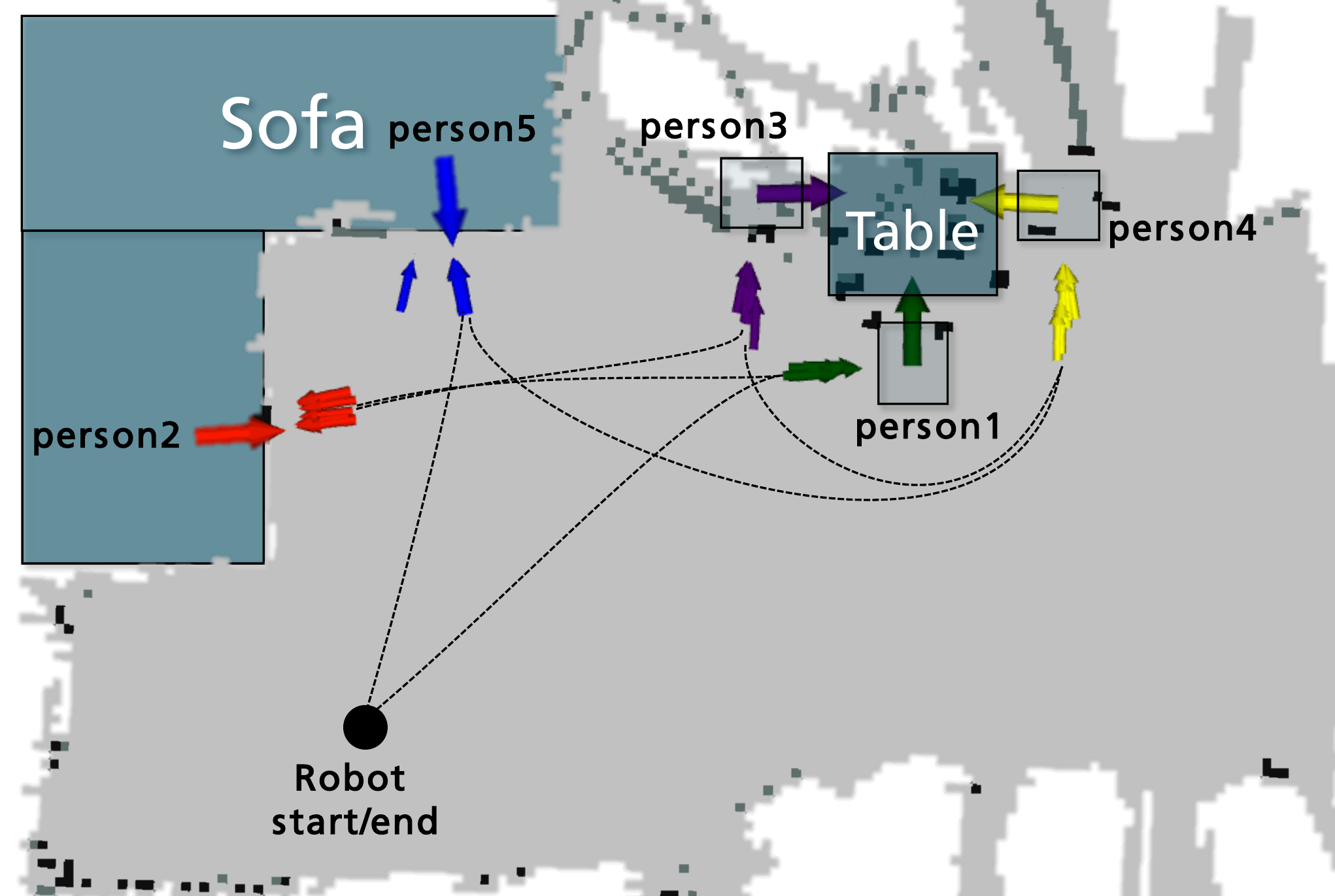}
            \caption{Gridmap with HRI poses}
            \label{fig:mobika-rviz}

\end{figure}

\begin{figure}[ht!]
    \centering
            \subfigure[Results Unidirectional Search]{\includegraphics[width=0.4\textwidth]{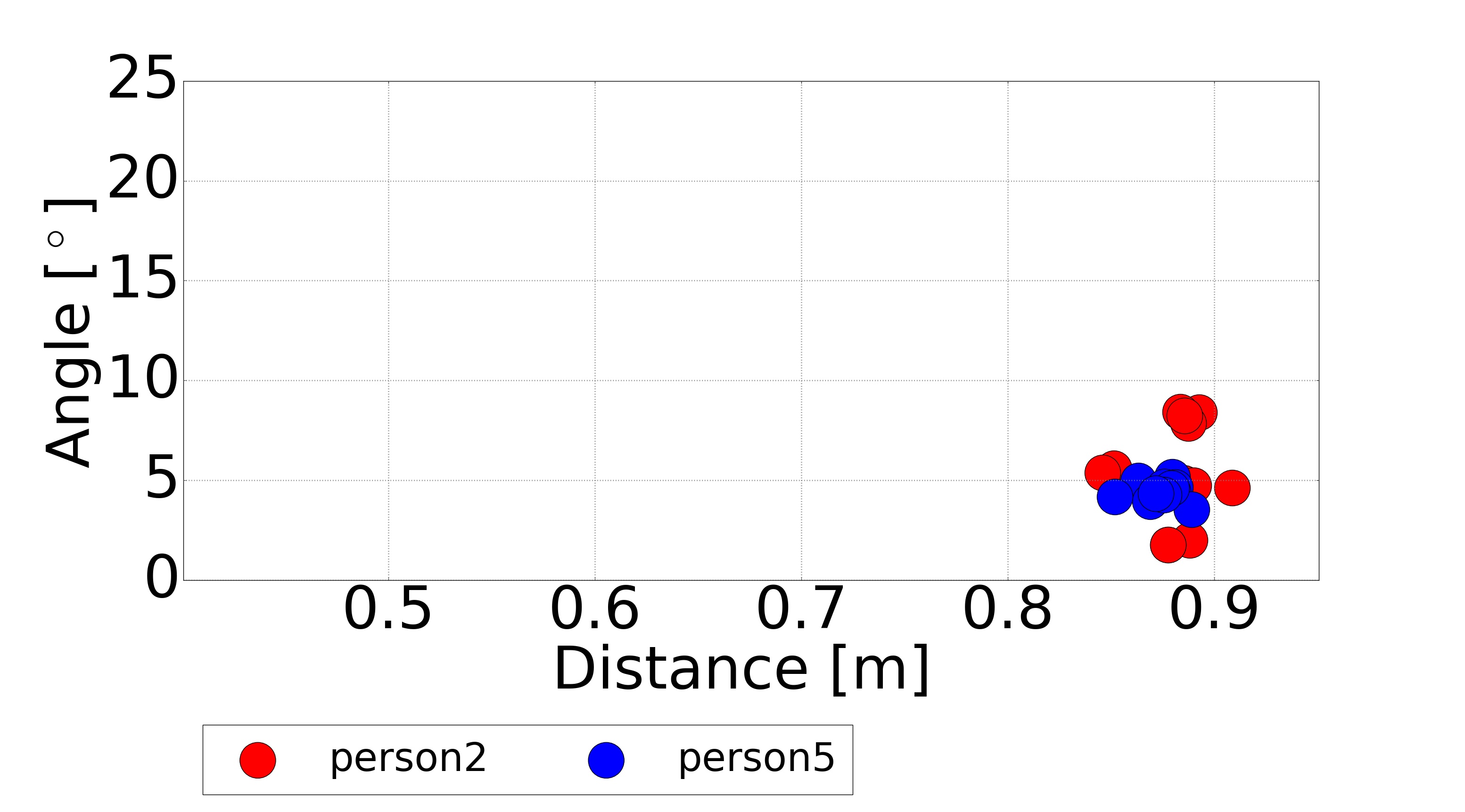}}
            \subfigure[Results Bidirectional Search]{ \includegraphics[width=0.39\textwidth]{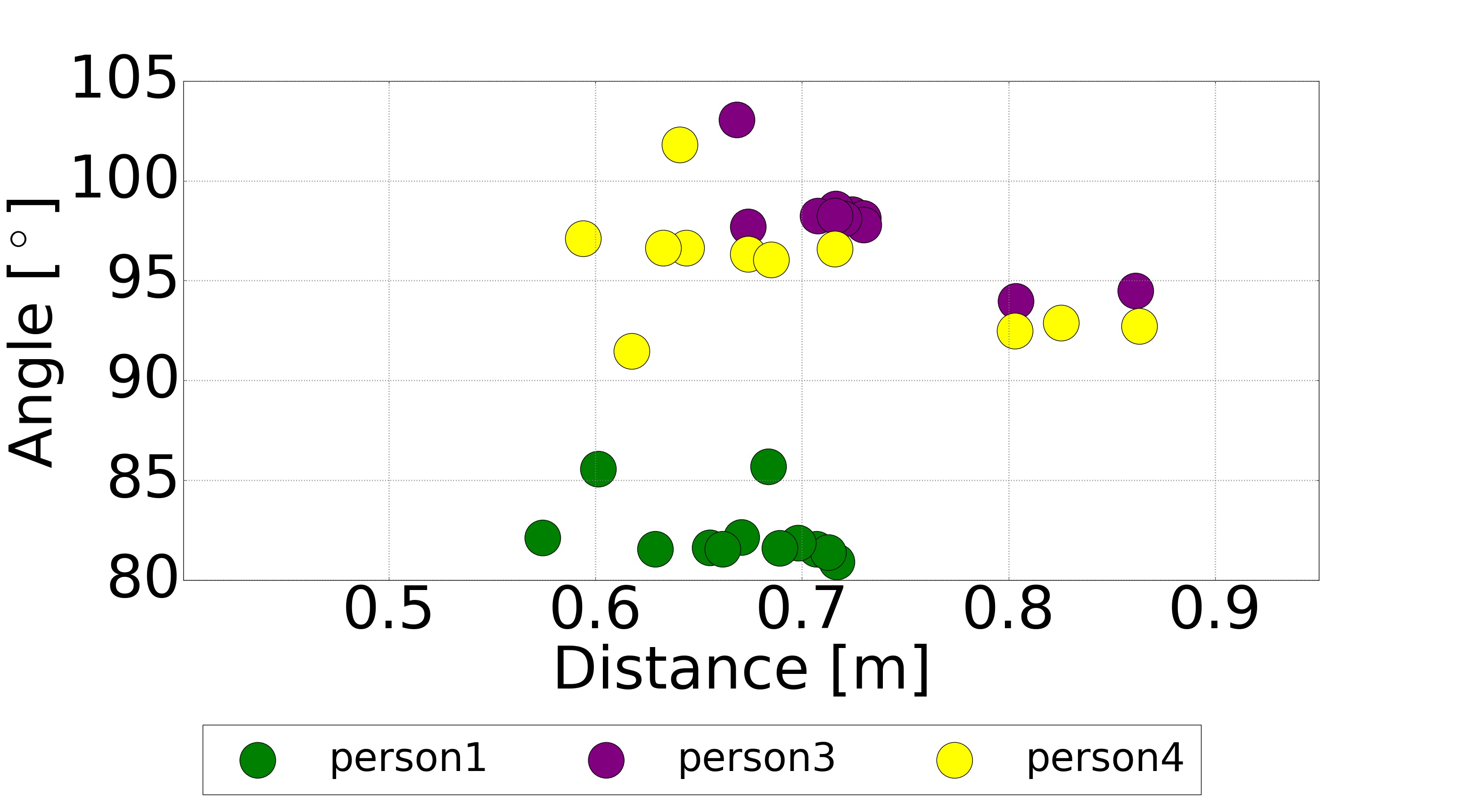}}
 
            \caption{Final Pose of the Robot in Relation to the Human}            
            \label{fig:results}

\end{figure}
 
\subsection{User-Feedback and Observations During Tests in an Elderly Care Home} \label{sec:test}
 
In addition to the lab tests, we tested the robot in a real life scenario. In the EmAsIn project,  a system consisting of three main components, namely a server, kinect sensors and MobiKa was used to activate eight residents with dementia in the group room, thus making everyday life more varied (Fig.~6). The portfolio of activations consisted of games, quizzes, picture galleries, and karaoke. 

For the evaluation, observations by the involved scientists were collected. In addtion, questionnaires from four elderly people and three care workers were evaluated. All respondents answered the question of whether the mobile robot platform is too human, negative. In addition also the speed of the robot and the approach behaviour was considert adequate by all respondents. Both the size and the shape of the mobile robot platform had a pleasant effect on the residents.
About  $\SI{85}{\%}$ of respondents said that the final pose for interaction stayed well within the social distance and was not to close.

\begin{figure}[ht!]            
  \centering
  \includegraphics[width=0.45\textwidth]{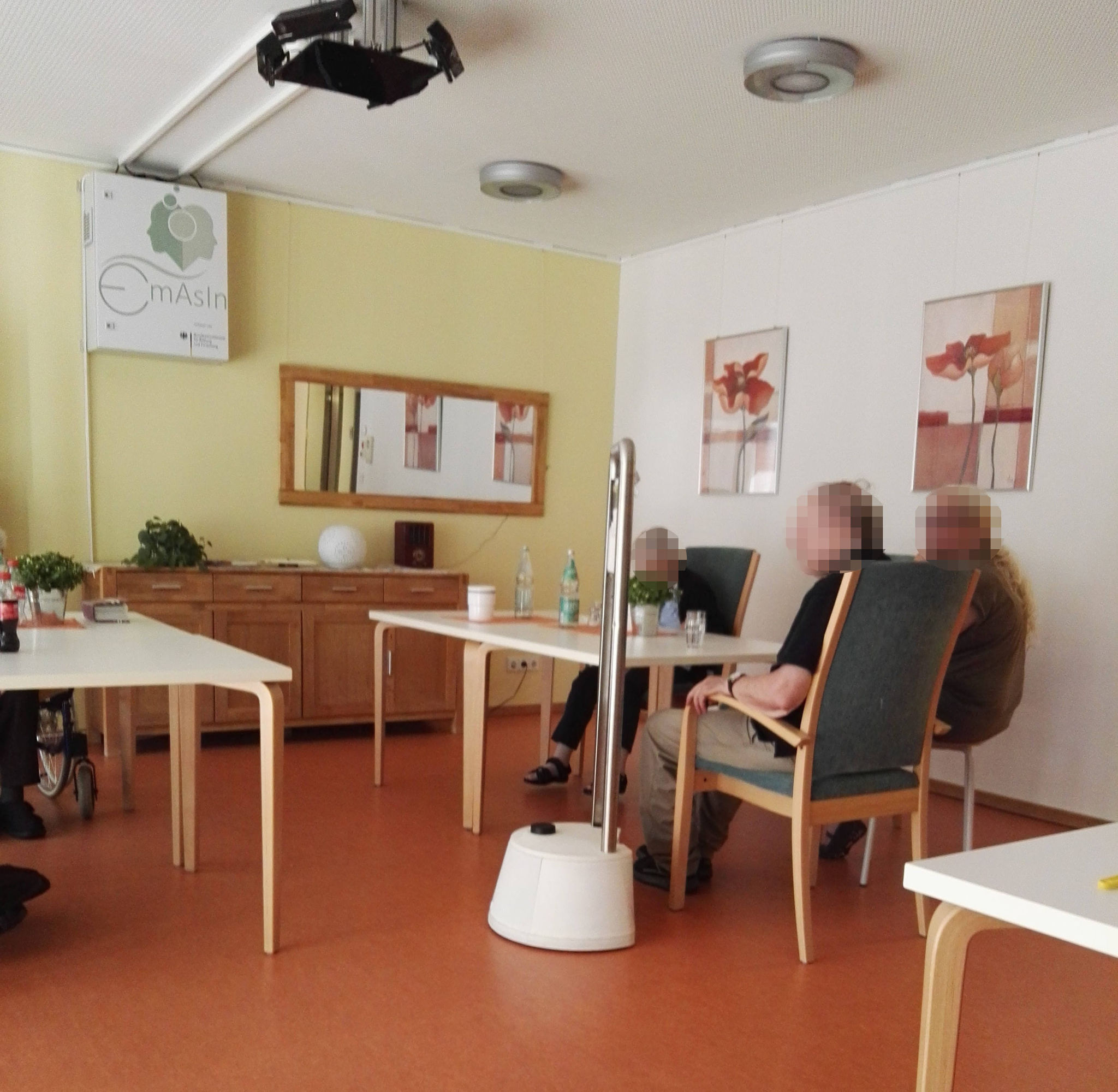}
  \caption{Interaction of MobiKa with elderly people}
  \label{fig:mobika-teningen}

\end{figure}

           
MobiKa successfully activated elderly people by approaching them. We observed and got the feedback that the robot behavior successfully maintains the human comfort, e.g. the robot approaching was a good trade-off between how close to move for the user-interaction without scaring persons. It was observed, that approaching the person already motivated them to interact with the robot. This is a big advantage compared to simple tablet solutions without the robot. The usage of robots, especially the touchscreen was new for most of the elerly people. Therefore, the users needed a short introduction from supervisors. Activities, e.g. quiz also activated nearby people that led to a group activity.

\bigbreak

%% file: sections/conclusion.tex
\section{CONCLUSIONS AND OUTLOOK}
\label{sec:conclusion}

The functional design of MobiKa enables versatile user interaction by using a height adjustable tablet which allows multimodal communication while standing, sitting and lying down (e.g. after a fall). In combination with low-cost components, the functional design helps to minimize the cost. 
MobiKa can navigate autonomously through a pre-mapped environment. For approaching the human, we developed an efficient and robust algorithm. The approaching was evaluated in laboratory tests, which indicated human-aware navigation. Additionally, it was tested in a care home to activate elderly people with dementia. The open infrastructure enables universal expansion options.

Due to the positive feedback and the economic potential of MobiKa, the development of MobiKa will continue in two directions. On the one hand, we will extend its functionalities. On the other hand, we will work on commercialization of our platform to make it available for end users. Moreover, we will include the approaching algorithm into the navigation stack to re-use it for other robots.

%% file: sections/acknowledgement.tex
\section{ACKNOWLEDGMENT}

This  work  has  received  funding  from  the  German Ministry of Education and Research (BMBF) under grant agreement No 16SV7362 for the EmAsIn project as well as funding from the European European Union’s  2020  research  and  innovation  programme  under the Marie Skodowska-Curie grant agreement No 721619 for the SOCRATES project.  